\pdfoutput=1
\documentclass[10pt,twocolumn,letterpaper]{article}

\usepackage[pagenumbers]{cvpr} 

\usepackage{lineno}
\usepackage{wrapfig}
\usepackage{multirow}
\usepackage{graphicx}
\usepackage{amsmath}
\usepackage{amssymb}
\usepackage{booktabs}

\usepackage{bm}
\usepackage{bbding}
\usepackage{color}
\usepackage{pifont}
\usepackage{csquotes}

\definecolor{cvprblue}{rgb}{0.21,0.49,0.74}
\usepackage[pagebackref,breaklinks,colorlinks,allcolors=cvprblue]{hyperref}
\usepackage[accsupp]{axessibility}  

\def\paperID{5892} 
\def\confName{CVPR}
\def\confYear{2025}

\title{CheXWorld: Exploring Image World Modeling \\for Radiograph Representation Learning}

\author{
Yang Yue$^1$\thanks{Equal contribution.\ \ \ \ \ \ \ \ \ \ \ \ \ \ \ \textsuperscript{\Envelope}Corresponding authors.} \ \ \ \ 
Yulin Wang$^1$$^*$\ \ \ \ 
Chenxin Tao$^1$ \ \ \ \ 
Pan Liu$^2$ \ \ \ \ 
Shiji Song$^1$ \ \ \ \ 
Gao Huang$^{1}$ \!\textsuperscript{\Envelope}\\
{\small$^1$Tsinghua University\ \ \ \ \ \ $^2$PLA General Hospital}\\[-0.5ex]
{\small\texttt{yueyang22@mails.tsinghua.edu.cn,\ gaohuang@tsinghua.edu.cn}}
}

\newcommand{\methodname}{CheXWorld}
\begin{document}
\maketitle
\begin{abstract}
     Humans can develop internal world models that encode common sense knowledge, telling them how the world works and predicting the consequences of their actions. 
     This concept has emerged as a promising direction for establishing general-purpose machine-learning models in recent preliminary works, {e.g.}, for visual representation learning. In this paper, we present {\methodname{}}, the first effort towards a self-supervised world model for radiographic images. Specifically, our work develops a unified framework that simultaneously models three aspects of medical knowledge essential for qualified radiologists, including 
     1) local anatomical structures describing the fine-grained characteristics of local tissues ({e.g.}, architectures, shapes, and textures); 
     2) global anatomical layouts describing the global organization of the human body ({e.g.}, layouts of organs and skeletons); 
     and 
     3) domain variations that encourage {\methodname{}} to model the transitions across different appearance domains of radiographs
     ({e.g.}, varying clarity, contrast, and exposure caused by collecting radiographs from different hospitals, devices, or patients).
     Empirically, we design tailored qualitative and quantitative analyses, revealing that \methodname{} successfully captures these three dimensions of medical knowledge.  
     Furthermore, transfer learning experiments across eight medical image classification and segmentation benchmarks
     showcase that \methodname{} significantly outperforms existing SSL methods and large-scale medical foundation models. 
     Code \& pre-trained models are available at \url{https://github.com/LeapLabTHU/CheXWorld}.

\end{abstract}
    
\section{Introduction}
\label{sec:intro}

Intelligent agents like humans 
learn extensive background knowledge about the world 
\cite{lecun2022path}. This common-sense information is embedded in the agents' internal models of the world, playing a pivotal role in their perception, learning, and decision-making processes by simulating the world's dynamics, telling them what is plausible or impossible, and predicting the outcomes of their actions. Consequently, these world models enable agents to acquire new concepts and skills with minimal demonstrations and trials \cite{sorscher2022neural}. 
Recent works \cite{ijepa, iwm} have preliminarily verified the effectiveness of establishing visual representation learning approaches through the lens of world modeling. Pre-trained world models can produce 
semantically rich embeddings and adapt to various downstream tasks with limited data.



Like many other research areas, the field of medical imaging is experiencing a paradigm shift from task-specific models to general-purpose foundation models \cite{moor2023foundation,huang2023visual,zhou2023foundation,gao2023synthetic}, which are pre-trained on massive data and expected to encode meaningful medical knowledge. The idea of world modeling offers an approach to training a medical foundation model by capturing common sense information (\textit{e.g.} human anatomy) from medical images, which is a promising yet under-explored research direction.

\begin{figure*}[t]
    \centering
    \includegraphics[width=0.95\textwidth]{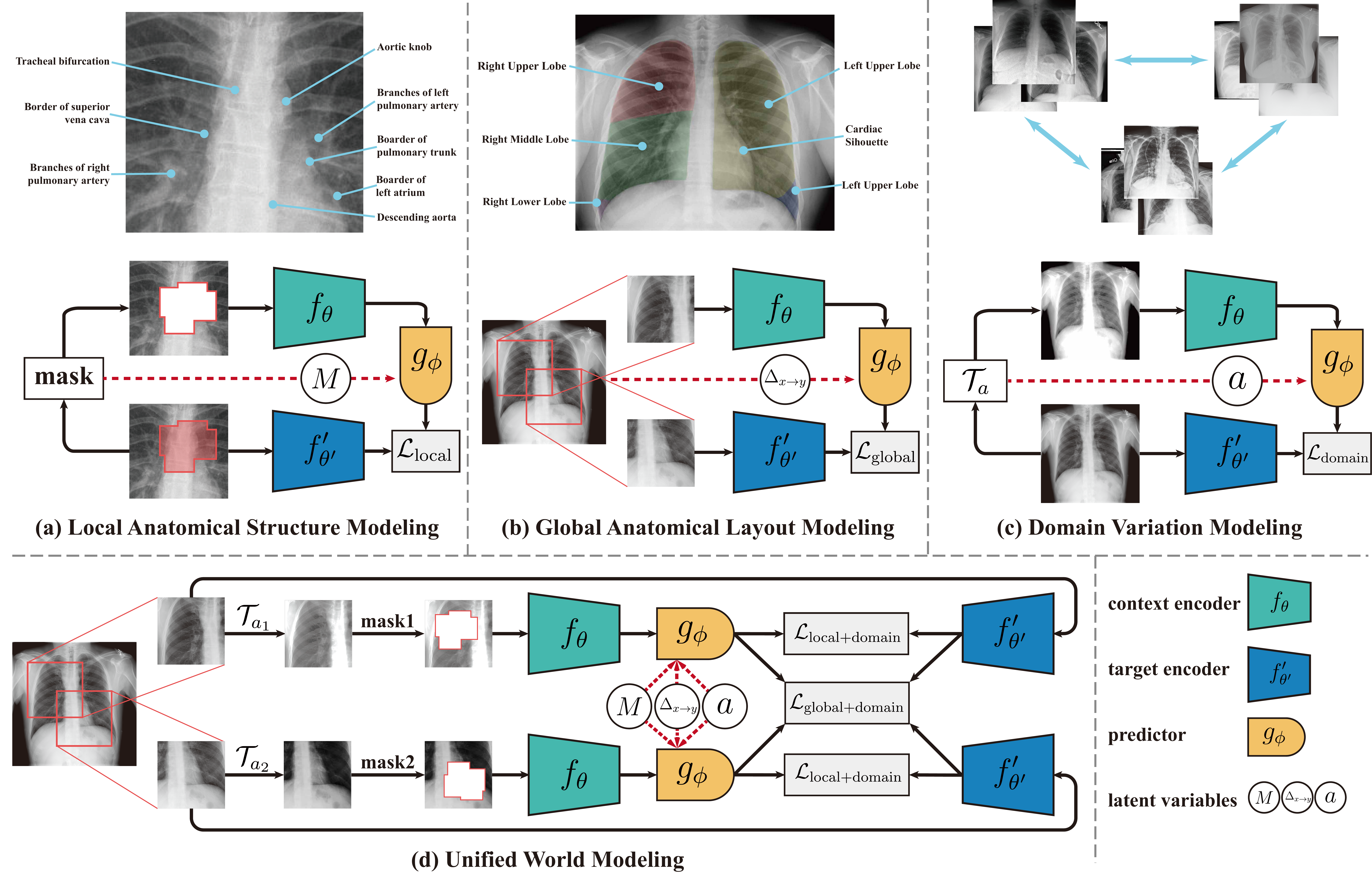}
    \vspace{-10pt}
    \caption{\textbf{Overview of the \methodname{} framework\protect\footnotemark.} The upper part of the figure depicts three dimensions of medical knowledge that are formulated in our framework, including (a) local anatomical structures describing the fine-grained characteristics of local tissues, (b) global anatomical layouts describing the global organization of the human body and (c) domain variations that encourage {\methodname{}} to model the transitions across different appearance domains of radiographs. The middle part of the figure illustrates the world modeling tasks corresponding to these aspects of medical knowledge. (d) shows our unified pipeline that combines the 
    merits of all three tasks.
    }
    \label{fig:method}
    \vspace{-17pt}
\end{figure*}

In this paper, we present \methodname{}, the first initiative toward world modeling for self-supervised representation learning on radiographic images. We propose and integrate three world modeling tasks that capture different dimensions of medical knowledge essential for qualified radiologists, depicted in Figure \ref{fig:method}. In particular, a radiologist with a strong understanding of human anatomy can identify different anatomical structures and determine the relative geometry between anatomical regions. Given the hierarchical nature of human anatomy, we introduce world modeling tasks that focus on both local and global levels of anatomy: \underline{\textit{local anatomical structure modeling}} aims to predict specific anatomical characteristics and structures within a localized anatomical region, such as bones, airways, blood vessels, and lung segments; \underline{\textit{global anatomical layout modeling}} learns the overall arrangement and spatial relationships of various anatomical regions within the human body, understanding the relative positioning of organs and tissues such as the heart, lungs, diaphragm, and rib cage. Additionally, medical images are typically gathered from multiple domains with notable appearance differences due to varying equipment and acquisition techniques \cite{imperfect}. Experienced radiologists can recognize the physical reality behind domains with different appearances and facilitate the diagnosis process. Motivated by this, we hypothesize that radiologists quickly adapt to diverse domain changes by learning sophisticated internal world models that are capable of simulating transitions between domains. 
Driven by our hypothesis, we introduce the \underline{\textit{domain variation modeling}} task, which aims to learn an expressive feature space that encodes predictable transitions between domains. 
Finally, we design an integrated pipeline that performs the three world modeling tasks simultaneously, effectively incorporating medical anatomical knowledge and achieving robustness across domains.

\footnotetext{The x-rays are taken from \href{https://www.kenhub.com/en/library/anatomy/medical-imaging-and-radiological-anatomy}{kenhub.com}, Radiopedia \cite{radiopedia58938} and ChestX-ray14 \cite{wang2017chestx}.}

It is noteworthy that the three aspects we consider, \emph{i.e.}, local anatomical structure modeling, global anatomical layout modeling, and domain variation modeling, represent core elements of radiographic expertise, each contributing to a comprehensive framework for capturing medical knowledge. While they may not encompass every dimension of medical understanding, they are fundamental to radiology and critical for radiograph representation learning, as extensively demonstrated in \cite{zhou2021models,zhou2023learning,guan2021domain,ouyang2022causality}.

Empirically, we validate that \methodname{} effectively captures the three dimensions of medical knowledge through a series of analytical experiments, including visualizations of predictor outputs with the help of a generative model and a domain variation sensitivity test.  Moreover, we extensively evaluate \methodname{} on eight medical image classification and segmentation benchmarks. Our model consistently outperforms competitive self-supervised learning baselines with comparable backbone capacity, pre-training data, and pre-training computational cost.

\section{Related Work}

\label{sec:related}
\textbf{World Modeling.}
In the machine learning community, world modeling originates from model-predictive control \cite{bryson2018applied,camacho2007constrained} and is a common practice in reinforcement learning, where it predicts the future state of the environment based on the agent's action \cite{ha2018recurrent,ha2018world,hafner2019dream,hafner2020mastering}. Recently, building general world models featuring profound comprehension of common sense knowledge is believed to be a crucial step towards general artificial intelligence \cite{lecun2022path,runwayml2023worldmodels}. Video generation models \cite{kang2024far} trained on vast data (\textit{e.g.} SoRA\cite{openai2024videogeneration} and VideoPoet \cite{kondratyuk2023videopoet}) are world models that perform predictions in input space. Another line of work makes predictions in the latent space \cite{hafner2023mastering,hu2023gaia,lecun2022path,ijepa,bardes2023v}, which more closely resembles the world model in the human brain. World modeling is also an effective approach for visual representation learning, where the model predicts missing parts 
of the visual inputs \cite{ijepa,bardes2023v,baevski2023efficient,baevski2022data2vec,iwm}.  
Our framework builds on this foundation, where our model acquires knowledge—such as human anatomy and domain variations—through learning to predict unobserved outcomes from specific actions.
Our key contribution lies in providing insights into modeling medical knowledge from radiographic images and designing a framework that unifies three world modeling tasks.


\textbf{Self-supervised learning in medical imaging.}
Self-supervised learning (SSL) \cite{mae,oquab2024dinov2,du2024probabilistic,wang2023efficienttrain} provides a practical solution to mitigating annotation scarcity in medical imaging.  A large body of works incorporate restorative SSL tasks, which typically involve reconstructing the original data from certain image corruptions, such as random masking \cite{selfmedmae, xiao2023delving, zhou2023foundation}, pixel shuffling \cite{zhou2021models}, and patch order shuffling \cite{tao2020revisiting, zhou2023learning, pang2022popar}. Another line of works \cite{sowrirajan2021moco, azizi2021big, zhou2020comparing, chaitanya2020contrastive, perez2024rad} adapt global or dense contrastive learning to medical imaging. 
TransVW\cite{haghighi2021transferable} and DiRA \cite{taher2022caid,haghighi2022dira} explore combining discriminative and restorative methods. 
Additionally, learning from anatomy is a common topic for medical SSL. SAM \cite{yan2022sam} enforces pixel-level anatomical correspondence between images, and Adam \cite{adamv2} learns part-whole correspondence that reflects the hierarchy of human anatomy. Recently, SSL-based foundation models trained on large-scale datasets gained significant attention, such as RETFound \cite{zhou2023foundation} for retinal imaging and UNI \cite{chen2024towards} for pathology.

Compared with existing works, \methodname{} is the first to introduce the concept of world modeling into medical SSL, capturing task-relevant medical knowledge through three tailored world modeling tasks. Unlike contrastive-based approaches such as Adam \cite{adamv2}, \methodname{} takes a technically distinct path by constructing equivariant \cite{dangovski2021equivariant,devillers2023equimod} representations—where input transformations result in predictable changes within the embedding space—rather than invariant representations. This approach allows \methodname{} to better model medical data without discarding valuable information. Moreover, our unified SSL framework delivers state-of-the-art performance across eight benchmarks, demonstrating the significant potential of designing SSL methods grounded in the philosophy of world modeling.

\section{Basic Framework of World Modeling}
\label{sec:basic}

\textbf{World modeling.}
We begin by briefly introducing the concept of \emph{world modeling} \cite{ha2018world,lecun2022path}, which forms the basis for our proposed method. The primary motivation behind \emph{world modeling} is to predict the unobserved parts $y$ of a world (\emph{e.g.}, a visual environment, an image, or a video) based on an observed context $x$. This prediction can be formulated across various dimensions, such as spatial (predicting the unseen regions of the data) and temporal (foreseeing the consequence of an action). In fact, the paradigm of \emph{world modeling} takes inspiration from human visual cognition. For example, when presented with the right upper lobe region of a chest radiograph as the context $x$, the internal world of a radiologist could imagine that there are two lobes (right middle and right lower) below the visible region and two lobes on the left 
, forming the target $y$.


\begin{wrapfigure}{r}{0.5\linewidth}

  \begin{center}
    \vspace{-4ex}
    \hspace{-17pt}
        \begin{minipage}{1.1\linewidth}
      \includegraphics[width=\linewidth]{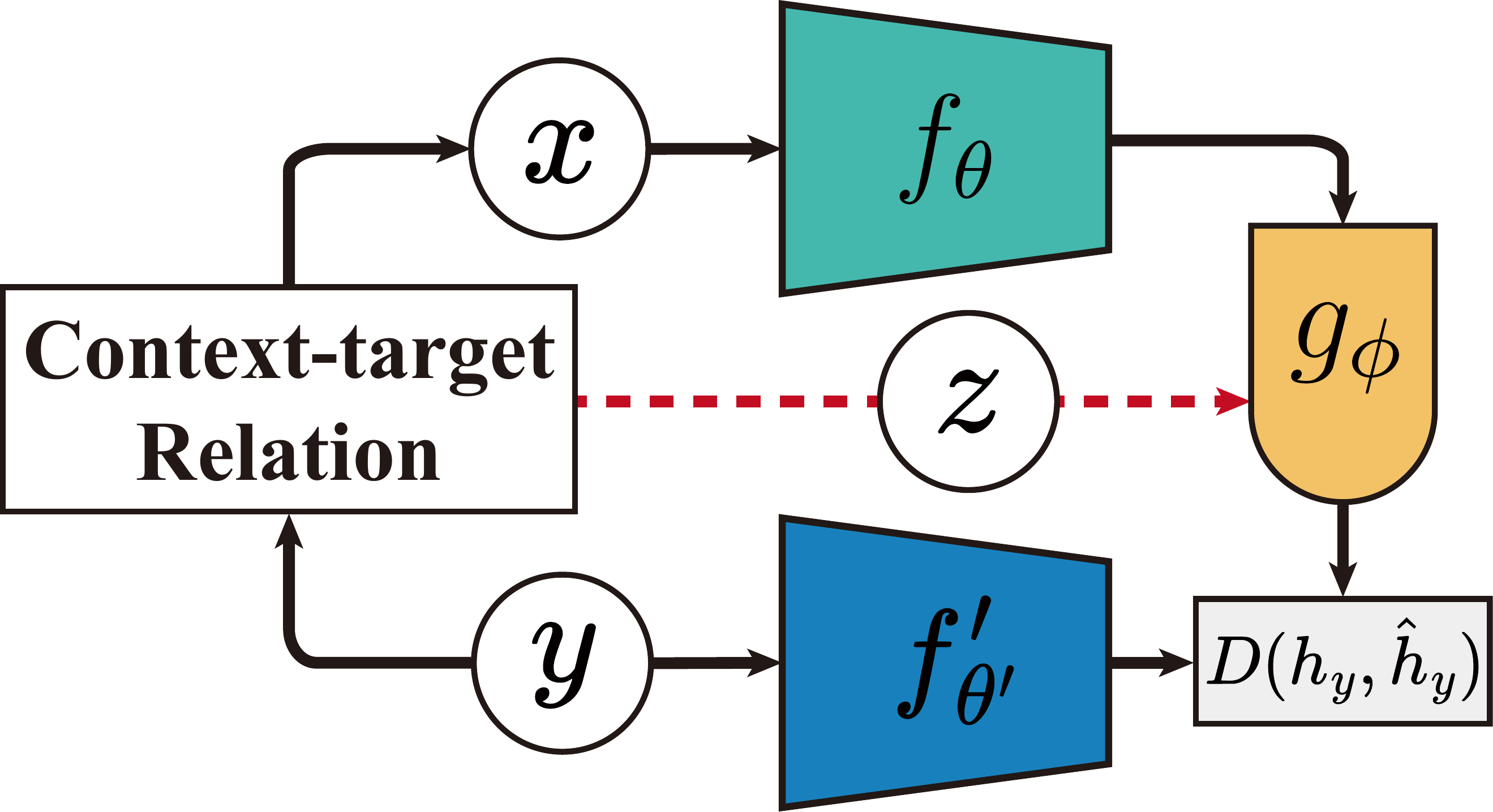}
    \vskip -0.1in
    \caption{\label{fig:basic}\textbf{A basic framework of world modeling.}}
    \end{minipage}
    \vspace{-25pt}
    \end{center}
\end{wrapfigure}
\textbf{A basic framework.}
Here, we introduce our basic framework of instantiating world modeling, upon which we will further discuss how to establish a radiology world model by formulating various medical knowledge in Section \ref{sec:main_method}. 
In specific, our basic framework mainly follows the joint-embedding predictive architecture (JEPA) \cite{lecun2022path, ijepa}, as shown in Figure \ref{fig:basic}. We consider two encoders: the \textit{context encoder} $f_\theta$ and the \textit{target encoder} $f'_{\theta'}$, which encode the context $x$ and target $y$ into representations $h_x$ and $h_y$, respectively. A \textit{predictor} $g_\phi$ learns to predict the target embedding $h_y$ using the context embedding $h_x$, conditioned on an additional latent variable $z$ that indicates how the target relates to the context. For example, $z$ can be the spatial locations of $y$ with respect to $x$, or an action that leads to the transition from context to target $x\rightarrow y$ (\emph{e.g.}, image transformations).
The latent variable $z$ carries the information that makes $y$ predictable based on $x$, effectively modeling the inherent variation and uncertainty of the real world. 
The primary objective of world modeling is:
\begin{equation}
\setlength\abovedisplayskip{5pt}
\setlength\belowdisplayskip{5pt}
    \textnormal{minimize} \quad D(h_y, \hat{h}_y) = D(h_y, g_\phi(h_x;z)),
\end{equation}
where $D(\cdot, \cdot)$ denotes the prediction error. Notably, the predictions are performed in the abstract representation space, contrasting with generative world models that 
predict every detail of the target. This mirrors an important characteristic of intelligent agents: information filtering, \emph{i.e.}, eliminating the irrelevant details in the data during the perception process. For example, a radiologist can 
infer the anatomical layout of a partly missed image, but it's nearly impossible for him to recover each pixel value of the unseen regions.

In implementation, we set the target encoder to be the exponential moving average of the context encoder following \cite{ijepa,bardes2023v,baevski2022data2vec}.
Moreover, we use the Vision Transformer (ViT) \cite{vit} architecture for the context/target encoder and the predictor. For the rest of this paper, the context and target feature $h_x, h_y$ are \textbf{sequences of patch embeddings} produced by the ViTs. To perform patch-level feature predictions, we attach mask tokens 
to the context features and feed the combined sequence to the predictor. The output representations of the mask tokens are used as the final prediction.

\section{\methodname{}: World Modeling for Radiograph Representation Learning}
\label{sec:main_method}

This section extends the idea of world modeling to radiology. We propose three world modeling tasks tailored for radiographs and establish a unified \methodname{} framework that seamlessly integrates them. Importantly, the three tasks are designed to model three critical dimensions of medical knowledge essential for qualified radiologists.
To be specific, these tasks are organized hierarchically from low-level to high-level, namely 1) \emph{local anatomical structure modeling} that aims to learn the fine-grained structures
of local anatomical regions, 2) \emph{global anatomical layout modeling} that seeks to learn the global geometry of the human body
(\emph{e.g.}, the layout of organs and skeletons), and 
3) \emph{domain variation modeling} that learns to model the transition across different appearance domains of radiographs. 
We first elaborate on the details of each task. Then, we present a unified framework that comprehensively combines the characteristics and merits of all three tasks, yielding a powerful foundation model that produces semantically rich and transferable representations.

\subsection{Local Anatomical Structure Modeling}
\label{section:_knowledge_local}

At a relatively low, fundamental level, various types of tissues 
(e.g., bones, muscles, and epithelial tissues)
form the structural organization of the human body. 
Understanding the fine-grained characteristics (\emph{e.g.}, shapes, sizes, appearance, and textures) of these micro-structures is essential knowledge that radiologists must possess. 
To build up a learning procedure that enables neural networks to acquire such local anatomical knowledge, we consider a mask-and-reconstruction task. The model predicts fine-grained details of a masked tissue region based on its peripheral, surrounding information, as shown in Figure \ref{fig:method}(a). This encourages the networks to develop an internal model that captures the intricate structures and local continuities of human microanatomy.
The large, continuous prediction target also prevents reliance on low-level features.


Specifically, the image mask is a union of four randomly selected rectangular regions 
of the image following \cite{ijepa}. Let $M$ denote the set of 2D patch locations that are masked from the context input. Given an input image (crop) $I$, the image patches with locations contained in $M$ are dropped from the image to create the context $x = \operatorname{Mask}(I, M)$, while the target is the entire image. The context and target are then fed to the corresponding encoders, producing representations $h_x=f_\theta(x)$ and $h_y=f'_{{\theta'}}(y)$. Note that the context encoder exclusively processes visible patches, following \cite{mae, ijepa}. In the predictor, the mask tokens $m$ carrying the positional encoding of the masked locations are concatenated with the context feature $h_x$, which forms the input token sequence to the predictor. The predictor output $\hat{h}_y$ is given by:
\begin{equation}
\setlength\abovedisplayskip{5pt}
\setlength\belowdisplayskip{5pt}
\begin{aligned}
        \hat{h}_y &= g_\phi(h_x; M)\\
        &= g_\phi\left(({h}_x + p_x)\oplus \left\{m+\operatorname{PE}(u,v)\ \right\}_{(u,v)\in M}\right),
\end{aligned}
\end{equation}
where $\oplus$ is the concatenation operation along sequence dimension, $p_x$ is the positional embedding corresponding to the context, $u$ and $v$ are image patch location indices along height and width dimensions, $\operatorname{PE}(\cdot)$ is the sinusoidal positional encoding proposed by \cite{vaswani2017attention}. The prediction loss is computed only on the masked locations, which is given by:
\begin{equation}\label{eq:local}
\setlength\abovedisplayskip{5pt}
\setlength\belowdisplayskip{5pt}
\mathcal{L}_{\operatorname{local}}(x,y) = \sum_{c\in M} \left\|g_\phi(f_\theta(x); M)_c - f'_{\theta'} (y)_c \right\|_2^2,
\end{equation}
where $c$ is the 2D image patch location that belongs to $M$.


\subsection{Global Anatomical Layout Modeling}


In addition to understanding local microstructures like tissues, it is also important for radiologists to possess comprehensive medical knowledge regarding how groups of tissues or systems of organs are anatomically organized and assembled into the full bodies of humans. 
Inspired by this, we develop the global anatomical layout modeling task (see Figure \ref{fig:method}(b)), seeking to predict the feature of an out-of-context area based on its relative position to a given context, thus facilitating understanding of the global structure of human bodies.
Notably, this idea is orthogonal to the technique proposed in section \ref{section:_knowledge_local} (see section \ref{sec:abl} for ablation studies) as it focuses on formulating the long-range topological relationships of different tissues or organs instead of the fine-grained anatomical structures within local tissues.

\begin{wrapfigure}{r}{0.5\linewidth}
  \begin{center}
    \vspace{-4ex}
    \hspace{-17pt}
        \begin{minipage}{1.1\linewidth}
      \includegraphics[width=\linewidth]{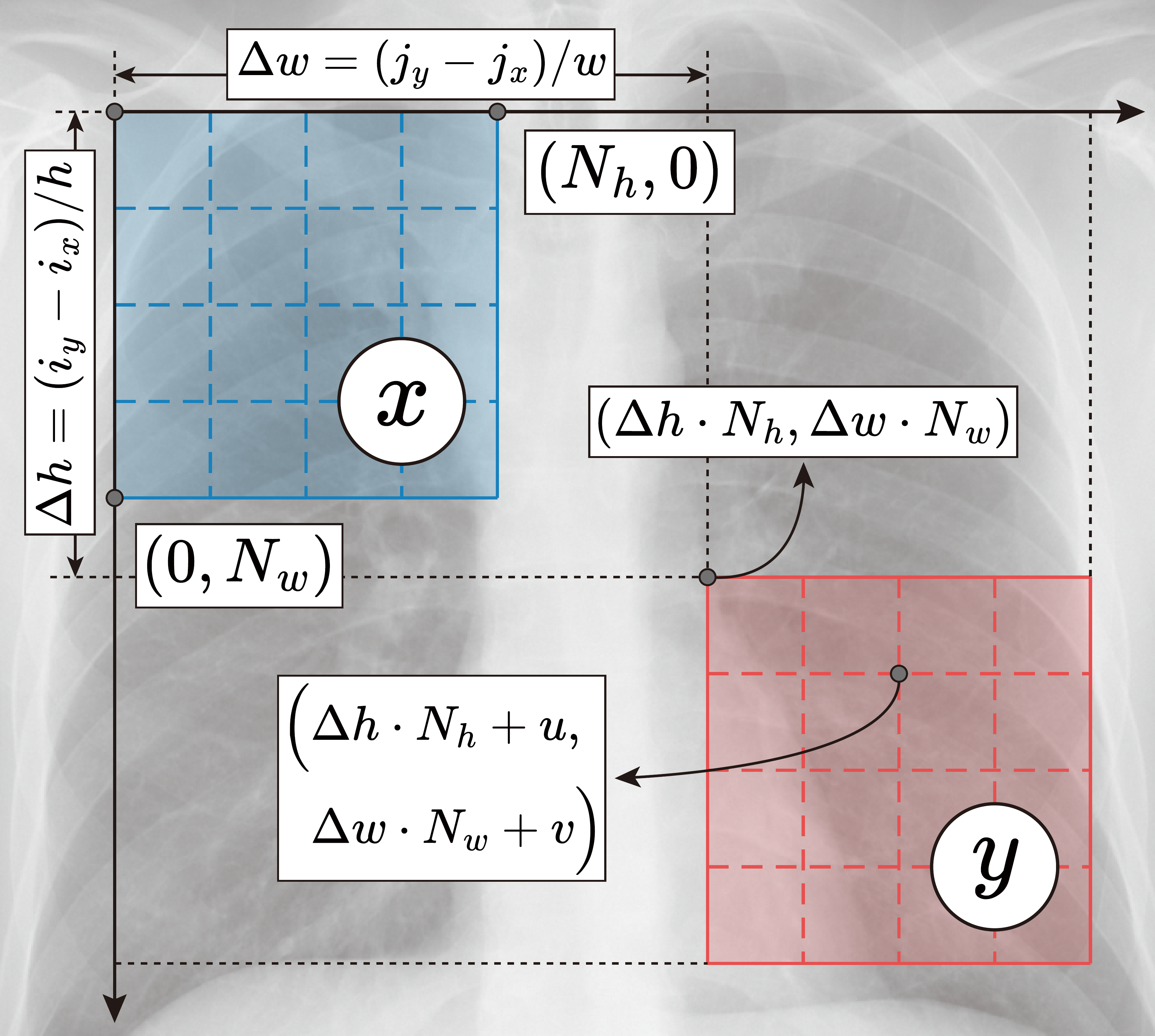}
    \vskip -0.1in
    \caption{\label{fig:rel_pos}\textbf{Formulation of global anatomical layout modeling.}}
    \end{minipage}
    \vspace{-20pt}
    \end{center}
\end{wrapfigure}
As illustrated in Figure \ref{fig:method}(b), we randomly crop two anatomical areas from the original radiograph, serving as context $x$ and target $y$, respectively. The world model learns to predict $y$ from $x$ given their relative position information $\Delta_{x\rightarrow y}$. In particular, we assume the context and target image crop share the same spatial size of $h$ pixels in height and $w$ pixels in width, with the left-top corner located at $(i_x, j_x)$ and $(i_y, j_y)$ in the pixel space. The relative position information is defined as the proportional relative displacement between the two regions:
\begin{equation}
\setlength\abovedisplayskip{5pt}
\setlength\belowdisplayskip{5pt}
    \Delta_{x\rightarrow y} = (\Delta h, \Delta w) = \left(\frac{i_y - i_x}{h}, \frac{j_y-j_x}{w}\right).
\end{equation}
To properly represent the location of the image patches in the target $y$, we establish a coordinate system over the context image patches, with the top-left corner patch located at the origin $(0, 0)$ and the bottom-right corner patch located at $(N_h-1, N_w-1)$, where $N_h, N_w$ are the number of tokens along the height and width dimensions. 
As demonstrated in Figure \ref{fig:rel_pos}, the coordinates of the image patch at the $u$-th row and $v$-th column
in the target image are given by:
\begin{equation}
\setlength\abovedisplayskip{5pt}
\setlength\belowdisplayskip{5pt}
\begin{aligned}
\hspace{-6pt}
    \phi_{x\rightarrow y}(u,v) &= \left(\Delta h\cdot N_h + u, \Delta w \cdot N_w + v\right) \\
    &\mathrm{where} \quad u\in[0, N_h-1], v\in[0, N_w-1].
\end{aligned}
\end{equation}
Essentially, $\phi_{x\rightarrow y}(u,v)$ describes the target's location from the context's perspective, enabling the model to predict the target based on the context. Sinusoidal position encoding $\operatorname{PE}(\cdot)$ is then applied to the relative coordinates $\phi_{x\rightarrow y}(u,v)$, and the mask tokens carrying the positional embeddings of all the target image patches are fed to the predictor. 
The predictor output is given by:
\begin{equation}
\setlength\abovedisplayskip{5pt}
\setlength\belowdisplayskip{5pt}
\label{eq:global}
\begin{aligned}
\hspace{-5pt}
\hat{h}_y &= g_\phi(h_x; \Delta_{x\rightarrow y}) \\
&= g_\phi\left((h_x + p_x)\oplus \left\{m+\operatorname{PE}(\phi_{x\rightarrow y}(u, v)) \right\}_{u,v} \right),\\
&\mathrm{where}\quad u\in[0, N_h-1], v\in[0, N_w-1],
\end{aligned}
\end{equation}
and the loss function is given by:
\begin{equation}
\setlength\abovedisplayskip{5pt}
\setlength\belowdisplayskip{5pt}
\mathcal{L}_{\operatorname{global}}(x, y) = \left\|g_\phi(f_\theta(x); \Delta_{x\rightarrow y}) - f_{\theta'}' (y)\right\|_2^2.
\end{equation}


\subsection{Domain Variation Modeling}

Thus far, we have discussed two levels of anatomical knowledge of human bodies. 
Beyond this, radiologists are usually capable of flexibly adapting to the specific characteristics of each radiograph and understanding them despite variations \cite{guan2021domain,ouyang2022causality,cui2022confidence}.
As shown in Figure \ref{fig:method}(c), medical images often come from different sources (\emph{e.g.}, hospitals, techniques, equipment) and show diverse appearances \cite{imperfect} (\emph{e.g}., clarity, contrast, exposure).
Nevertheless, experienced radiologists can imagine the physical reality behind the radiographs with varying appearances, which facilitates the diagnosis process. We hypothesize that radiologists' internal world models can simulate the variations across domains, thus obtaining an objective view of the scanned body.


Inspired by this, we propose a domain variation modeling task that learns how image features change across domains, enabling cross-domain adaptability. We simulate domain shifts using data augmentation and construct context-target pairs to model these transitions.
The model is tasked with predicting the feature outcomes of the inverse effect of an image transformation determined by certain augmentation parameters, as shown in Figure \ref{fig:method}(c). The learned representation space is designed to be equivariant \cite{dangovski2021equivariant,devillers2023equimod},
\emph{i.e.}, input transformation leads to predictable output change, 
which preserves sufficient information across domains, mimicking radiologists' internal models.

Specifically, the target $y$ is an augmented version of the original image. Another transform parameterized by $k$ scalars (denoted by $a \in \mathbb R^k$) is further applied to the target to produce the context $x = \mathcal T_a(y)$, where $\mathcal T$ is an augmentation pipeline consisting of Gaussian blur, brightness, contrast, and gamma adjustments. The parameter $a$ contains the strength and other configurations of the augmentation. The world model learns to model the feature transformation from the context to the target domain (\emph{i.e.} the inverse effect of $\mathcal T_a$) conditioned on the parameter $a$, which can be regarded as the \enquote{action} of domain transition. The prediction is also performed with mask tokens to be compatible with the other two world modeling tasks. The parameter $a$ is combined with the mask tokens via a lightweight policy network $\pi$. The forward pass and loss function of domain variation modeling are given by:
\begin{equation}
\setlength\abovedisplayskip{5pt}
\setlength\belowdisplayskip{5pt}
\begin{aligned}
    m_{a, p} &= \pi(m+p, a), \\
    \hat{h}_y &= g_\phi(h_x; a) \\
    &= g_\phi\left((h_x + p_x)\oplus \left\{m_{a, \operatorname{PE}(u,v)} \right\}_{u,v} \right), \\
\hspace{-5pt}\mathcal{L}_{\operatorname{domain}}(x, y) &= \left\|g_\phi(f_\theta(x); a) - f_{\theta'}' (y)\right\|_2^2,
    \label{eq:appr}
\end{aligned}
\end{equation}
where $p$ is the positional embedding of an image patch, $m_{a, p}$ denotes the mask token carrying both spatial and domain transition information. 

\begin{figure*}[t]
    \centering
    \includegraphics[width=0.8\textwidth]{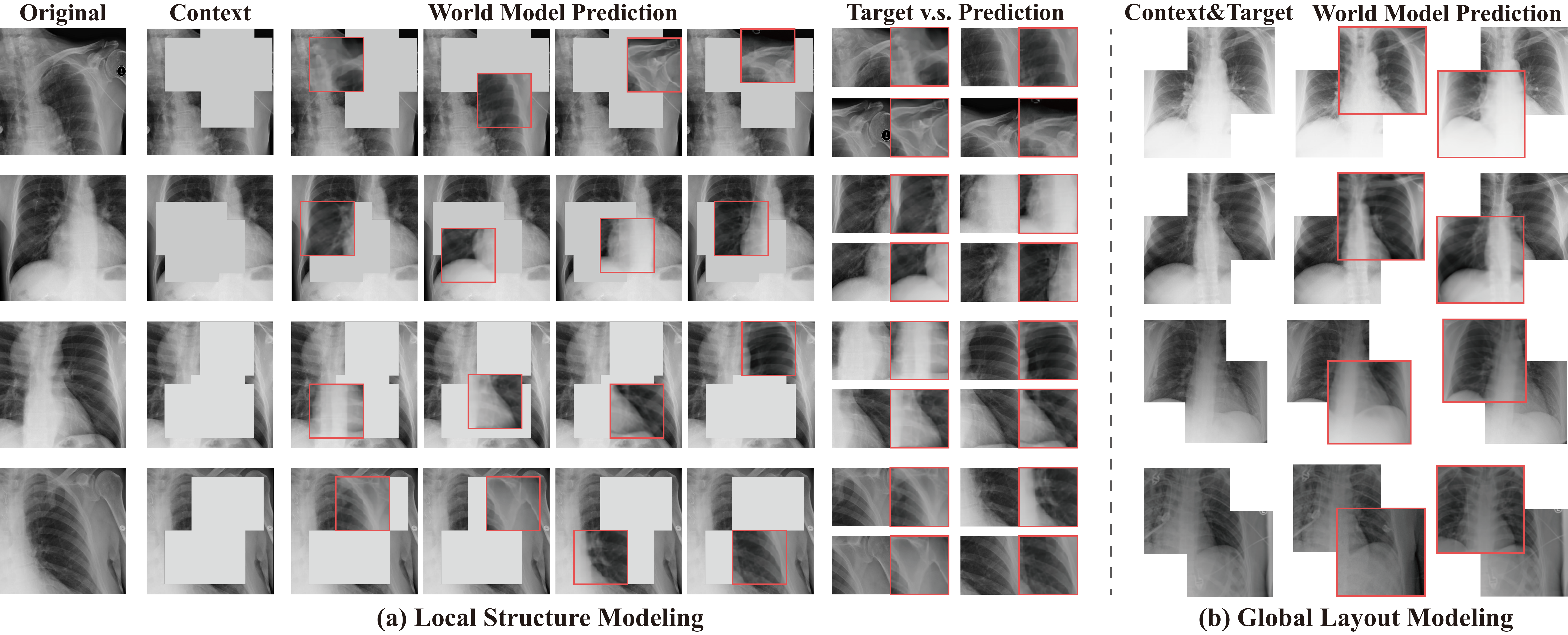}
    \vspace{-10pt}
    \caption{\textbf{Visualization of the \methodname{} predictor outputs} (zooming in for details). The images presented in this figure were not included in the pre-training of \methodname{} or the training of the diffusion model.
    Regions in red bounding boxes denote the predictor outputs that are mapped to pixel space using the RCDM \cite{bordes2021high} framework.
    In (a), gray areas indicate masked regions excluded from the context. In (b), the two overlapping regions alternately serve as context and target. 
    }
    \label{fig:rcdm}
\end{figure*}
\begin{figure*}[t]
    \vskip -10pt
    \begin{center}
    \begin{minipage}{0.9\columnwidth}
    \centering
    \includegraphics[width=0.75\columnwidth]{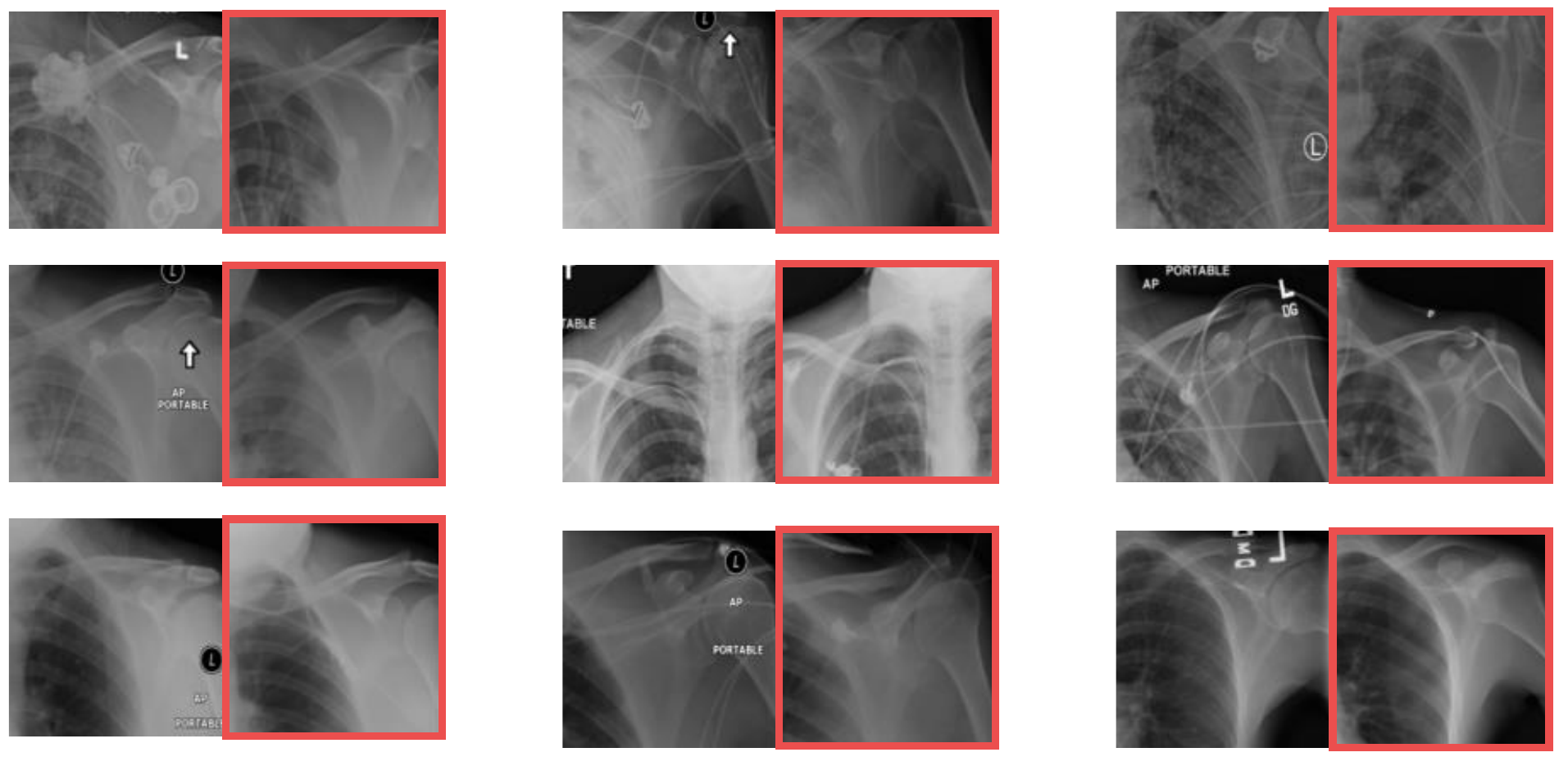}
            \vskip -0.08in
            \caption{\textbf{\methodname{} prioritizes relevant medical features over spurious signals} (\textit{e.g.}, lateral markers) in the image. Regions in red boxes denote the predictor outputs.}
            \label{fig:artifact}
        \end{minipage}
    \hspace{0.05in}
    \begin{minipage}{1.15\columnwidth}
    \centering
\includegraphics[width=0.85\columnwidth]{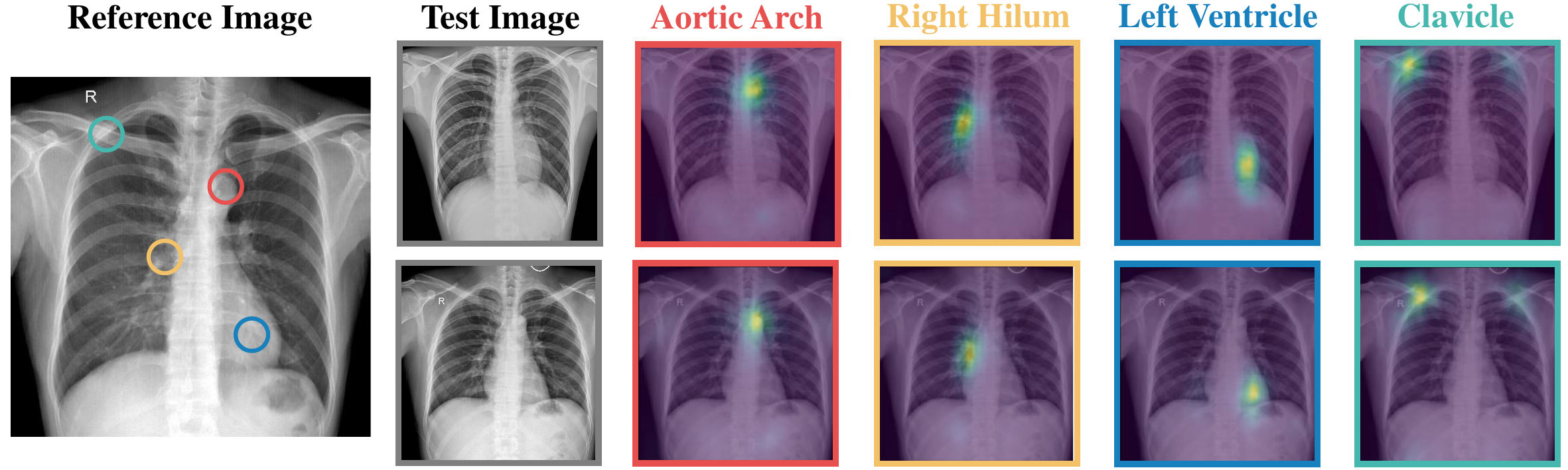}	
        \vskip -0.1in
        \caption{\textbf{\methodname{} learns anatomical correspondence.}  We compute pixel-level embeddings using RoI-pooling \cite{girshick2014rich} and calculate the embedding similarity between four anatomical landmarks in a reference image and each pixel in the test image to create feature similarity heatmaps.}
    \label{fig:keypoint}
    \end{minipage}
    \end{center}
    \vskip -0.3in
 \end{figure*}

\subsection{The Unified Framework}

Having established the individual tasks of local anatomical structure modeling, global anatomical layout modeling, and domain variation modeling, we now 
integrate these tasks into a cohesive system. This integration is important for creating a comprehensive model that can leverage multiple dimensions of medical knowledge simultaneously and maximize the combined benefits of each task.



Instead of simply adding three loss functions together, a more elegant solution is to design unified contexts, targets, and latent variables to perform the three world modeling tasks in a single forward pass. As illustrated in Figure \ref{fig:method}(d), the training pipeline of \methodname{} starts with sampling two different image crops from the image $y_1, y_2$, which serve as the targets. The targets are then augmented and masked with two different configurations $(a_1, M_1)$ and $(a_2, M_2)$, producing the contexts $x_1$, $x_2$:
 \begin{equation}
\setlength\abovedisplayskip{5pt}
\setlength\belowdisplayskip{5pt}
x_i = \operatorname{Mask}(\mathcal T_{a_i}(y_i), M_i), \, i=1,2.
\end{equation}
The contexts $x_1, x_2$ and targets $y_1, y_2$ are then encoded into representations $h_{x_1}, h_{x_2}, h_{y_1}, h_{y_2}$. Note that the targets $h_{y_1}, h_{y_2}$ are predictable from the contexts $h_{x_1}, h_{x_2}$ given the proper conditions. Specifically, since $x_1$ is obtained by applying augmentation $\mathcal T_{a_1}$ and mask $M_1$ to $y_1$, predicting $y_1$ from $x_1$ with conditions $(a_1, M_1)$ simultaneously performs local anatomical structure modeling and domain variation modeling:
\begin{equation}
\setlength\abovedisplayskip{5pt}
\setlength\belowdisplayskip{5pt}
\label{eq:local_domain}
\begin{aligned}
    \mathcal L_{1\rightarrow 1} &= \mathcal L_{\operatorname{local+domain}}(x_1, y_1) \\
    &=  \sum_{c\in M_1} \left\|g_\phi(h_{x_1}; M_1, a_1)_c - f_{\theta'}' (y_1)_c \right\|_2^2.
\end{aligned}
\end{equation}
Moreover, $x_1$ can be derived from $y_2$ by moving the input window by $-\Delta_{x_1\rightarrow y_2}$ and applying augmentation $\mathcal T_{a_1}$, hence the predictor conditioning on $(\Delta_{x_1\rightarrow y_2}, a_1)$ performs global anatomical structure modeling and domain variation modeling at the same time.
\begin{equation}
\label{eq:global_domain}
\setlength\abovedisplayskip{5pt}
\setlength\belowdisplayskip{5pt}
\begin{aligned}
    \mathcal L_{1\rightarrow 2} &= \mathcal L_{\operatorname{global+domain}}(x_1, y_2) \\
    &= \left\|g_\phi(h_{x_1}; \Delta_{x_1\rightarrow y_2}, a_1) - f_{\theta'}' (y_2) \right\|_2^2.
\end{aligned}
\end{equation}
Similarly, $\mathcal L_{2\rightarrow 2}, \mathcal L_{2\rightarrow 1}$ can be computed between $(x_2, y_2)$ and $(x_2, y_1)$, respectively. Note that the context features and target features are reused twice when computing the loss \eqref{eq:local_domain} and \eqref{eq:global_domain}. Finally, the model is trained with the combination of $2\times2=4$ supervisions computed from the two pairs of context and target.
\begin{equation}
\setlength\abovedisplayskip{5pt}
\setlength\belowdisplayskip{5pt}
    \mathcal L = \mathcal L_{1\rightarrow 1} + \mathcal L_{2\rightarrow 2} + \mathcal L_{1\rightarrow 2} + \mathcal L_{2\rightarrow 1}.
\end{equation}
With this unified objective, \methodname{} showcases deep comprehension of human anatomy in different scales and pixel intensity distributions across various domains.


\section{Experiments}

\begin{table*}[t]
    \centering
    \caption{\textbf{Performance on five downstream classification benchmarks.} Accuracy is reported for RSNA, and the area under the ROC curve (AUROC) is reported for the rest of the datasets. The best two results are \textbf{bold-faced} and \underline{underlined}, respectively. 
    }
    \label{tab:result_main}
    \vspace{-5pt}
    \resizebox{0.95\linewidth}{!}{
    \centering
    \begin{tabular}{l|l|l|ccccc}
    \toprule
        Method & Pre-train Data & Backbone  & VinDr-CXR & ShenZhen& ChestX-ray14 & RSNA (CLS) & CheXpert \\
        \midrule
        Scratch & - & ViT-B  & 70.22$\pm$1.95 & 82.24$\pm$0.60 & 71.69$\pm$0.32 & 66.59$\pm$0.39 & 80.78$\pm$0.03\\
        \midrule
        MoCo-v3 \cite{mocov3} & IN-1k (1.3M) & ViT-B  & 87.25$\pm$0.63 &  92.85$\pm$1.00 & 79.20$\pm$0.30 & 72.79$\pm$0.52 & 87.12$\pm$0.36 \\ 
        DINO \cite{dino} & IN-1k (1.3M) & ViT-B  & 82.89$\pm$1.10 & 90.39$\pm$4.29 & 78.37$\pm$0.47 & 71.27$\pm$0.45 &  87.01$\pm$0.62 \\ 
        BEiT \cite{bao2021beit}& IN-21k (12M) & ViT-B & 85.93$\pm$1.98 & 92.87$\pm$1.08 & 79.91$\pm$0.24 & 72.78$\pm$0.37 & 87.77$\pm$0.38\\
        \midrule
        LVM-Med \cite{mh2024lvm}& Medical (1.3M) & ViT-B  & 88.22$\pm$0.44 & 94.81$\pm$1.32 & 80.08$\pm$0.09 & 72.75$\pm$0.44 & 88.07$\pm$0.25 \\
        MAE \cite{mae, xiao2023delving}& X-rays (0.5M) & ViT-B   & 92.76$\pm$0.18& 97.63$\pm$0.21& 83.0$^\dagger$ & 73.75$\pm$0.24 & \underline{89.3}$^\dagger$ \\[0.4ex]
        \multirow{2}{*}{SimMIM \cite{ma2022benchmarking, simmim}} & IN-21k (12M) \&  & \multirow{2}{*}{Swin-B}&  \multirow{2}{*}{\underline{92.81$\pm$0.31}}  & \multirow{2}{*}{\underline{98.09$\pm$0.13}} & \multirow{2}{*}{83.04$\pm$0.15} & \multirow{2}{*}{\underline{74.09$\pm$0.39}} & \multirow{2}{*}{89.14$\pm$0.22} \\[-0.6ex]
        & X-rays (0.9M) & & & & & \\[0.4ex]
        \multirow{2}{*}{Adam-v2 \cite{adamv2}}& IN-21k (12M) \&  & \multirow{2}{*}{ConvNeXt-B}  & \multirow{2}{*}{91.46$\pm$0.33} & \multirow{2}{*}{97.80$^\dagger$} & \multirow{2}{*}{\underline{83.4}$^\dagger$} & \multirow{2}{*}{73.40$\pm$0.88} & \multirow{2}{*}{88.90$\pm$0.36} \\[-0.6ex]
        & X-rays (0.9M) & & & & & \\[0.4ex]
        
                \midrule
        \methodname{} & X-rays (0.5M) & ViT-B  & \textbf{95.24$\pm$0.13} & \textbf{98.88$\pm$0.06} & \textbf{83.58$\pm$0.05} & \textbf{75.03$\pm$0.39} & \textbf{89.63$\pm$0.13}\\
            \midrule
        \multirow{2}{*}{\color{gray} Rad-DINO$^\ddag$ \cite{perez2024rad}} & \small{\color{gray} LVD (142M) \&}  & \multirow{2}{*}{\color{gray} ViT-B} & \multirow{2}{*}{\color{gray}95.16$\pm$0.16} & \multirow{2}{*}{\color{gray}98.20$\pm$0.17} & \multirow{2}{*}{\color{gray}83.61$\pm$0.08} & \multirow{2}{*}{\color{gray}74.51$\pm$0.46} & \multirow{2}{*}{\color{gray} 88.94$\pm$0.15} \\[-0.6ex]
        & \small{\color{gray} X-rays (public+private)} & & & & & \\
        \bottomrule
    \end{tabular}}
    \raggedright
    \emph{\footnotesize Results on ImageNet pre-trained models are adopted from \cite{ma2022benchmarking}. $^\dagger$Results reported by the original authors \cite{mae,adamv2}. $^\ddag$Rad-DINO \cite{perez2024rad} is included for reference but not for direct comparison, as it requires 2560 GPU-hours (20 times more than ours) and is trained on both public and private datasets.}
    \vspace{-10pt}
\end{table*}

In this section, we first demonstrate \methodname{}'s world modeling capability with a series of analytical experiments. Then, we empirically compare \methodname{} with other self-supervised methods as baselines on eight downstream tasks.
Finally, we provide ablation studies to highlight the effect of each building block of our method.

\textbf{Implementation details.}  
\methodname{} is pre-trained on $\sim$0.5M  frontal chest X-rays from several public datasets.
The visual backbone is a ViT-Base \cite{vit}. The model is trained for 300 epochs, taking 16 hours on 8 RTX 4090 GPUs.
Please refer to the appendix for more details.

\subsection{\methodname{} is a Strong World Predictor}
\label{sec:vis}

\textbf{Visualization of anatomical modeling.}
To better understand the predictions made by \methodname{}, we train a generative model that maps the average-pooled predictor output back to pixel space following the RCDM framework \cite{bordes2021high}. Specifically, we construct context-target pairs in the same way as in the local and global anatomical modeling tasks. Then, we feed the context to a frozen pre-trained \methodname{} model to obtain the model's feature prediction, on which a diffusion model conditions to produce the target image in pixel space. 
As shown in Figure \ref{fig:rcdm}(a), the \methodname{} predictor perfectly recovers the appearance of bones and tissues within various masked areas, and Figure \ref{fig:rcdm}(b) shows that the model makes global-level predictions that are consistent with the context.
Further visualizations also show that \methodname{} filters artifacts in the image (Figure \ref{fig:artifact}) and learns anatomical correspondence (Figure \ref{fig:keypoint}).

\textbf{Awareness to domain variations.}
We design a domain sensitivity test to verify the model's ability to capture the uncertainty in a radiograph's appearance attributes. Specifically, we first generate a candidate set of targets $\{y_i\}_{i=1}^n$ by applying augmentations with different configurations to the same image.
Then, we construct a set of context-target-latent triplets $\{(x_i, y_i, a_i)\}$ by applying another augmentation $x_i=\mathcal T_{a_i}(y_i)$. The model is asked to predict the target $y_i$ given the context $x_i$ and latent $a_i$. For each predicted output $\hat{y}_i$, we calculate the top-k recall rate of the true target $y_i$ over the entire candidate set using L2 distance\footnote{In our experiments the candidate set size is fixed at $n=64$. This process is repeated across multiple images, and the final result is obtained by averaging the outcomes.}. 
As shown in Table \ref{tab:domain_recall}, our model achieves an average top-5 recall of 77.67 (10 times higher than random choice), demonstrating the model's strong discriminative ability across domains. We also observe a significant drop in the recall rate when removing the domain condition $a$, indicating that the model indeed makes predictions conditioned on $a$.

\subsection{\methodname{} Produces Transferable Representations}
\label{sec:results}

\begin{figure*}[t]
    \begin{center}
    \begin{minipage}{1.3\columnwidth}
\centering
\captionof{table}{\textbf{Results on segmentation (left) and few-shot learning (right) tasks.} The dice score and the AUROC score are reported for the segmentation and few-shot learning benchmarks respectively.}
\label{tab:result2}
\vspace{-5pt}
    \resizebox{\linewidth}{!}{
    \begin{tabular}{c|cc|ccc}
    \toprule
    \multirow{2}{*}{Method} & \multicolumn{2}{c|}{Segmentation} & \multicolumn{3}{c}{ Few-shot (MedFMC-ChestDR)} \\
       & SIIM-ACR & RSNA (SEG) & 1-shot & 5-shot & 10-shot \\
        \midrule
        LVM-Med \cite{mh2024lvm} & 82.19$\pm$0.30&  78.47$\pm$0.24& 57.55$\pm$0.81& 66.95$\pm$0.60&67.44$\pm$0.71 \\
        MAE \cite{mae, xiao2023delving}& 83.01$\pm$0.39& 77.39$\pm$0.25& \underline{61.31$\pm$0.27} & \underline{74.03$\pm$0.30}& \underline{75.26$\pm$0.36}\\
        SimMIM \cite{ma2022benchmarking, simmim}& 82.89$\pm$0.07 & 78.31$\pm$0.41 & 60.64$\pm$1.92 & 72.14$\pm$0.54 & 74.42$\pm$0.38\\
        Adam-v2 \cite{adamv2}& \underline{83.60$\pm$0.33} &  \underline{78.53$\pm$0.19} & 59.16$\pm$1.03 & 70.28$\pm$0.39 & 70.67$\pm$0.71\\
        \midrule
        \methodname{}& \textbf{84.58$\pm$0.34}& \textbf{79.02$\pm$0.25}& \textbf{64.60$\pm$1.00} & \textbf{75.19$\pm$0.51} &\textbf{76.40$\pm$0.25}\\
        \bottomrule
    \end{tabular}}
    \end{minipage}
    \hspace{0.05in}
    \begin{minipage}{0.6\columnwidth}
    \centering
    \includegraphics[width=0.95\columnwidth]{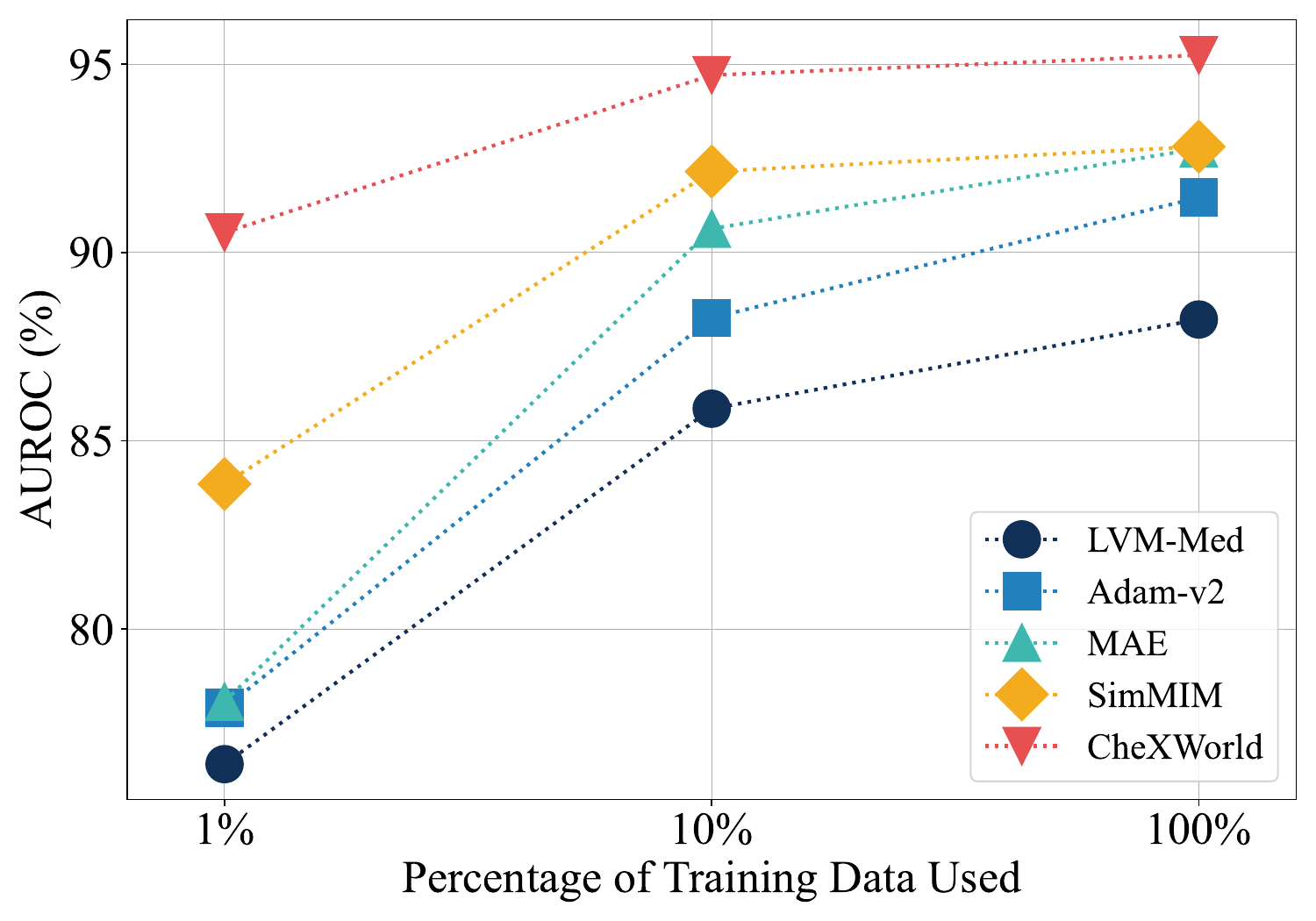}
            \vskip -0.1in
            \caption{Fine-tuning with 1\%, 10\%, and 100\% training data on VinDr-CXR.}
            \label{fig:vindr_frac}
        \end{minipage}
    \end{center}
    \vskip -0.2in
 \end{figure*}


\begin{figure*}[t]
    \begin{center}
    \begin{minipage}{0.7\columnwidth}
    \centering
    \includegraphics[width=0.9\linewidth]{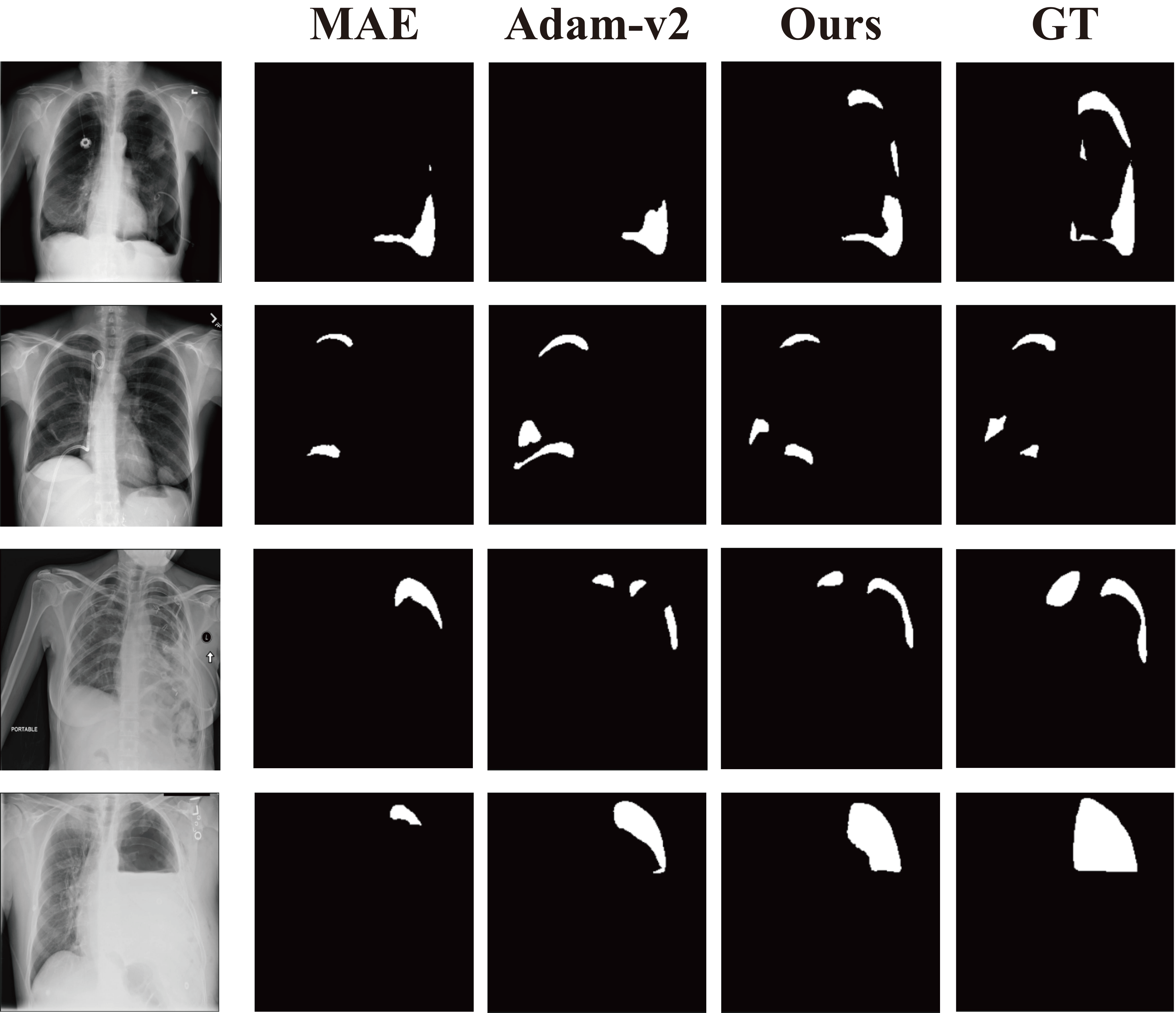}
    \vspace{-5pt}
    \caption{\textbf{Visualization of segmentation masks} 
    on SIIM-ACR pneumothorax dataset. GT stands for the ground truth masks.}
    \label{fig:seg_vis}
    \end{minipage}    
    \begin{minipage}{1.35\columnwidth}
    \centering
    \captionof{table}{\textbf{Ablation studies of the world modeling tasks and latent variables.} We report the results on VinDr-CXR and RSNA with 1\% proportion of training data. AUROC and accuracy are reported for the respective datasets. 
    \enquote{Local.}, \enquote{Global.}, and \enquote{Domain.} are abbreviations for local anatomical structure modeling, global anatomical layout modeling, and domain variation modeling, respectively. \enquote{Unified} indicates whether our unified formulation is applied, as opposed to a naive combination of losses.}
    \label{tab:result_abl}
    \vspace{-5pt}
    \resizebox{0.85\linewidth}{!}{
    \begin{tabular}{ccc|cc|c|cc}
    \toprule
        \multicolumn{3}{c|}{World Modeling Task} & \multicolumn{2}{c|}{Latent Variables} & 
        \multirow{2}{*}{Unified}&\multirow{2}{*}{VinDr 1\%} & \multirow{2}{*}{RSNA 1\%}\\
      Local. & Global. & Domain. & $\Delta_{x\rightarrow y}$ & $a$ &  & \\
      \midrule
    \checkmark & & & & & N.A. & 84.71$\pm$1.00& 65.89$\pm$0.54\\
       \checkmark & \checkmark& & \checkmark& & \checkmark& 88.31$\pm$0.88& 67.47$\pm$0.51\\
       \checkmark & & \checkmark & &\checkmark &  \checkmark&89.73$\pm$0.50& 68.27$\pm$1.00\\

       \checkmark & \checkmark& \checkmark& \checkmark& \checkmark& \checkmark& \textbf{90.53$\pm$1.01} & \textbf{70.01$\pm$0.33}\\
       \midrule
        \checkmark & \checkmark& \checkmark& \checkmark & \checkmark&  & 89.12$\pm$0.76 & 68.37$\pm$0.50\\
        \midrule
        \checkmark & \checkmark& \checkmark& \checkmark & & \checkmark& 89.91$\pm$0.60 & 68.98$\pm$0.60\\
        \checkmark & \checkmark& \checkmark& & \checkmark& \checkmark&87.60$\pm$0.72 & 66.56$\pm$0.39\\
        \bottomrule
    \end{tabular}}
    \end{minipage}
    \end{center}
    \vskip -0.3in
 \end{figure*}


\textbf{Setup and baselines.} We compare \methodname{} with several competitive baselines across eight classification and segmentation tasks using the following datasets: VinDr-CXR \cite{nguyen2022vindr}, ShenZhen-CXR \cite{shenzhen}, NIH ChestX-ray14 \cite{wang2017chestx}, CheXpert \cite{irvin2019chexpert}, MedFMC-ChestDR \cite{fmc}, SIIM-ACR Pneumothorax \cite{siim} and RSNA Pneumonia \cite{rsna}.


\begin{table}[h]
    \centering
    \vspace{-4pt}
    \caption{\textbf{Effectiveness of domain variation modeling.}}
    \vspace{-8pt}
    \label{tab:domain_recall}
    \resizebox{0.9\columnwidth}{!}{
    \begin{tabular}{c|ccc}
    \toprule
        Method &  Recall@1 & Recall@3 & Recall@5\\
        \midrule
        Random Choice & 1.56 & 4.68&7.81 \\
        w/o domain condition & 15.87 & 33.12& 44.37 \\
        \methodname{} & \textbf{38.49} & \textbf{65.90}& \textbf{77.67}\\
        \bottomrule
    \end{tabular}
    }
    \vspace{-18pt}
\end{table}

We attach task-specific heads (linear heads and U-Net decoders) and perform full fine-tuning on the pre-trained \methodname{} encoder. 
The input image resolution is fixed to $224^2$.
We compare \methodname{} with several self-supervised learning methods designed for general domain images \cite{mocov3, dino,bao2021beit,mae,simmim} and medical images \cite{adamv2, mh2024lvm} using base-scale backbones \cite{vit,liu2021swin,liu2022convnet}.
We use the radiology version of these methods \cite{xiao2023delving, ma2022benchmarking, perez2024rad} if available.


\textbf{Classification results} are shown in Table \ref{tab:result_main}. Our method consistently outperforms the best alternatives on all the benchmarks. 
Although SimMIM and Adam-v2 are initialized from ImageNet weights and pre-trained on a large dataset (0.9M X-rays) and,
yet still lag behind \methodname{} with considerable margins. For instance,  \methodname{} surpasses Adam-v2 by 4.04 in AUROC on VinDr-CXR and SimMIM by 0.54 on ChestX-ray14.
Our approach performs comparably to Rad-DINO across all benchmarks, despite Rad-DINO requiring 20 times more computational resources and utilizing private training data.
Additionally, one can observe a significant contrast in performance between models trained on photometric images (ImageNet) and models trained on medical images, highlighting the importance of in-domain transfer.

\textbf{Segmentation results} are presented in Table \ref{tab:result2} and Figure \ref{fig:seg_vis}. \methodname{} surpasses the leading baselines in terms of dice score and produces more accurate segmentation masks,
demonstrating its superior capability in both instance-level and dense prediction settings.

\textbf{Few-shot learning results} on the MedFMC dataset are presented in Table \ref{tab:result2}. \methodname{} demonstrates a clear advantage on the 1-shot learning task and consistently outperforms the second-best methods in both 5-shot and 10-shot scenarios, showcasing quick adaptation to new concepts by leveraging knowledge acquired through world modeling. 

\textbf{Sample efficiency results.}\footnote{Detailed numerical results can be found in the appendix.} The superiority of our method is further highlighted by reducing the training data proportion. As shown in Figure \ref{fig:vindr_frac}, we conduct experiments on VinDr-CXR with 1\%, 10\%, and 100\% of the training data. Notably, 
our method with 10\% of the data significantly outperforms all baselines trained on the entire dataset, reducing the annotation cost by over 90 percent.

\vspace{-2pt}
\subsection{Ablation Study}
\label{sec:abl}
\vspace{-2pt}


\textbf{Effectiveness of the world modeling tasks} is validated in the upper part of Table \ref{tab:result_abl}. Combining local anatomical structure modeling with either of the other two world modeling tasks yields significant performance gains. The best performance is achieved when all three world modeling tasks are integrated within our unified framework, indicating that these tasks reflect different aspects of medical knowledge that are effectively captured by \methodname{}.

\textbf{Role of the latent variables.} 
To verify that the latent variables (such as relative position $\Delta_{x\rightarrow y}$ for global anatomical layout modeling and transformation parameter $a$ for domain variation modeling) contribute to transfer learning performance, we test the models pre-trained without certain latent variables, shown in the bottom part of Table \ref{tab:result_abl}. Domain variation modeling without the condition $a$ degrades into a blind image restoration task, which yields inferior results compared with the original method. Removing the condition $\Delta_{x\rightarrow y}$ from global anatomical layout modeling leads to a sharp decrease in performance because it confounds the model with erroneous target spatial locations.

\section{Conclusion}

In this paper, we proposed \methodname{}, a self-supervised world modeling framework for radiographic images that simultaneously encodes local and global anatomical knowledge and appearance domain variations. By integrating three world modeling tasks into a unified pipeline, our model effectively captures the spatial and domain uncertainties of the image world, as verified by qualitative and quantitative analytical experiments.  Extensive transfer learning experiments on eight medical image analysis benchmarks demonstrate the state-of-the-art performance of our model. Our work offers new insights into representing and extracting knowledge from medical images, paving the way toward a general-purpose foundation model for medical vision.

\section*{Acknowledgements.} The work is supported in part by the National Key R\&D Program of China under Grant 2024YFB4708200 and National Natural Science Foundation of China under Grant
62321005.

{
    \small
    \bibliographystyle{ieeenat_fullname}
    \bibliography{main}
}

\clearpage
\setcounter{page}{1}
\maketitlesupplementary

\appendix

\section{Datasets and Baselines}
\subsection{Datasets}
\label{app:dataset}
Our experiments are based on eight chest X-ray datasets, including MIMIC-CXR \cite{johnson2019mimic} for pre-training; CheXpert \cite{irvin2019chexpert} and NIH ChestX-ray14 \cite{wang2017chestx} for both pre-training and fine-tuning; VinDr-CXR \cite{nguyen2022vindr}, ShenZhen-CXR \cite{shenzhen}, RSNA Pneumonia \cite{rsna}, MedFMC-ChestDR \cite{fmc}, and SIIM-ACR Pneumothorax \cite{siim} for fine-tuning. Detailed information on these datasets is provided below.

\begin{itemize}
    \item \textbf{MIMIC-CXR} \cite{johnson2019mimic} is one of the largest X-ray datasets, containing over 370k radiograph images from over 220,000 patient studies with paired radiology reports. We gather non-lateral scans from this dataset (about 230k images) and use this dataset for self-supervised pre-training.

    \item \textbf{CheXpert} \cite{irvin2019chexpert} contains about 218k images with 14 disease labels automatically extracted from radiology reports. We use this dataset for pre-training and conduct multi-label classification experiments on five conditions: atelectasis, cardiomegaly, consolidation, edema, and effusion. We report the performance on the official validation set (200 patients) with a held-off subset from the training set for model selection. The mean AUROC score over the five classes is reported for this dataset.

    \item \textbf{NIH ChestX-ray14} \cite{wang2017chestx} contains about 112k frontal-view chest radiographs, with annotations on 14 thoracic diseases: atelectasis, cardiomegaly, consolidation, edema, effusion, emphysema, fibrosis, hernia, infiltration, mass, nodule, pleural thickening, pneumonia, and pneumothorax. We use the training split of this dataset for pre-training and conduct disease classification experiments on the 14 classes. We follow the official split with 86k images for training and 25k for testing. The mean AUROC score over the 14 classes is reported for this dataset.

    \item \textbf{VinDr-CXR} contains 18,000 radiographs with expert annotations. Each radiograph is associated with 22 local findings and 6 global findings. We consider the multi-label classification task on the 6 global labels, including lung tumor, pneumonia, tuberculosis, other diseases, COPD, and no finding. We adopt the official split with 15,000 images for training and 3,000 images for testing. The mean AUROC score over the 6 classes is reported for this dataset.

    \item \textbf{ShenZhen-CXR} defines a binary classification problem where each radiograph is labeled with the presence of tuberculosis. We follow the data split provided by \cite{ma2022benchmarking} with the train/val/test split containing 463/65/134 images, respectively. The AUROC score is reported for this dataset.
    
    \item \textbf{RSNA Pneumonia} \cite{rsna} consists of over 26k radiographs, each categorized into one of three classes: normal, lung opacity, or no opacity but not normal. Additionally, expert-annotated bounding boxes highlight areas of lung opacity. This dataset is used for both classification and segmentation tasks. For classification, we frame it as a three-class problem, reporting top-1 accuracy. For segmentation, the bounding boxes are converted into segmentation masks and the mean dice score is reported.
    We follow the data split provided by \cite{ma2022benchmarking} with train/val/test split containing 21295/2680/2709 images, respectively. 

    \item \textbf{MedFMC-ChestDR} \cite{fmc} is a dataset tailored for few-shot adaptation. Each radiograph is associated with 19 common thoracic disease labels. The official competition consists of 1-shot, 5-shot, and 10-shot tracks, each with five different train/val splits. To ensure consistency, we use the first split in each track and report the mean performance averaged over five random seeds. The mean AUROC score is reported over the 19 classes for this dataset.

    \item \textbf{SIIM-ACR Pneumothorax} \cite{siim} comprises 12,047 radiographs with pixel-level annotations for pneumothorax. We perform binary segmentation on this dataset, with the mean dice score reported as the evaluation metric.
\end{itemize}

\subsection{Baselines}
We compare \methodname{} with several self-supervised learning methods developed for general-domain and medical images, including MoCo-v3 \cite{mocov3}, DINO \cite{dino}, BEiT \cite{bao2021beit}, MAE \cite{mae, xiao2023delving}, SimMIM \cite{simmim, ma2022benchmarking}, LVM-Med \cite{mh2024lvm}, Adam-v2 \cite{adamv2}, and Rad-DINO \cite{perez2024rad}. When possible, we leverage radiology-specific adaptations of these methods. For a fair comparison, all methods utilize models of comparable sizes, such as ViT-B \cite{vit}, Swin-B \cite{liu2021swin}, and ConvNeXt-B \cite{liu2022convnet}. Below, we provide a brief overview of each approach:

\begin{itemize}
    \item \textbf{MoCo-v3} \cite{mocov3} is a contrastive learning framework that employs a momentum encoder to create a dynamic dictionary for stable and effective representation learning. It explores additional training techniques to optimize vision transformer performance.
    \item \textbf{DINO} \cite{dino} pre-trains vision transformers with a self-distillation objective. Techniques like distribution centering and sharpening are incorporated to stabilize the training process.
    \item \textbf{BEiT} \cite{bao2021beit} is a masked image modeling (MIM) approach inspired by masked language modeling in natural language processing. The model predicts masked token indices generated by discrete variational autoencoders.
    \item \textbf{MAE} \cite{mae} is an encoder-decoder framework for MIM, predicting raw pixel values for masked patches. Only visible patches are passed to the encoder to improve computational efficiency. We use its radiology-adapted version introduced by \cite{xiao2023delving}.
    \item \textbf{SimMIM} \cite{simmim} is another MIM approach based on the Swin Transformer \cite{liu2021swin}. It employs random masking with a moderately large patch size and uses a simple linear decoder head. The radiology-adapted version from \cite{ma2022benchmarking} is used in our experiments.
   \item \textbf{LVM-Med} \cite{mh2024lvm} leverages a graph-matching formulation for contrastive learning, building a versatile model that integrates diverse medical image modalities and datasets.
   \item \textbf{Adam-v2} \cite{adamv2} focuses on learning anatomical structures in X-ray images hierarchically, using pre-training objectives that promote localizability, composability, and decomposability.
   \item \textbf{Rad-DINO} \cite{perez2024rad} extends DINOv2 \cite{oquab2024dinov2} by performing continuous pre-training on radiology datasets.
\end{itemize}

\section{Implementation Details}
\subsection{Pre-training}

\textbf{Data.} \methodname{} is pre-trained on the combination of three datasets: MIMIC-CXR \cite{johnson2019mimic}, NIH ChestX-ray14 \cite{wang2017chestx}, and CheXpert \cite{irvin2019chexpert} (following \cite{xiao2023delving}). We only use the frontal scans for pre-training, resulting in $\sim$0.5M radiographs in total. We exclude the validation/test split of the NIH Chest-Xray14 and CheXpert datasets from the pre-train dataset to avoid data leakage to the downstream tasks. 

\textbf{Architecture and optimization.} The context encoder is a ViT-Base with a patch size of $16\times 16$. The target encoder is the exponential moving average of the context encoder with an initial ratio equal to $0.996$ that gradually increases to $1.0$ following a cosine schedule. The predictor is 6 layers deep with 384-dimensional embeddings. We use sinusoidal functions \cite{vaswani2017attention} to encode the image patch positions following \cite{mae}. We use the AdamW optimizer \cite{loshchilov2018fixing} with $\beta_1=0.9, \beta_2=0.999$ with an initial learning rate of $2\times 10^{-4}$ and weight decay set to $0.05$. Gradient clipping is set to $1.0$ throughout our experiments. The learning rate schedule follows linear warmup for 40 epochs and cosine annealing afterward. L2 loss is computed between the raw predictor outputs and the layer-normalized target encoder outputs. The model is trained from scratch for 300 epochs with a batch size of 2048, 
taking 16 hours on a machine with 8 RTX 4090 GPUs, each with 24 GB memory.

\textbf{Local anatomical structure modeling.} We adopt a block-wise masking strategy \cite{ijepa}. The image mask is the union of four rectangular blocks with the scale $(0.15,0.2)$. We further shrink the context's visible area by a maximal factor of 0.25, which we found beneficial. The context encoder only processes unmasked patches, while the entire image takes the entire image as input. In the predictor, mask tokens corresponding to the masked locations are padded to the context. The loss is computed on masked locations.

\textbf{Global anatomical structure modeling.} We sample two random crops with the same spatial size with their scales in  $(0.3, 1.0)$ and aspect ratios in $(0.75, 1.33)$. The relative position information $\Delta_{x\rightarrow y}$ is obtained in pixel space and then used to determine the location of target image patches in the context's coordinate system. Note that the sinusoidal encoding function $\operatorname{PE}(\cdot)$ supports fractional inputs. Thus, the target patch locations $\phi_{x\rightarrow y}(u,v)$ can be encoded in the same way as the context patch locations. We compute prediction loss on all target patches.

\textbf{Domain variation modeling.} We simulate domain transitions with a set of augmentations, including brightness, contrast, gamma transform, and Gaussian blur. Given an input image $I$ (or an image crop), the target is obtained by applying brightness and contrast adjustment to the original image. Then, we apply another augmentation consisting of bright, contrast, gamma transform, and Gaussian blur, with the configurations of the augmentation stored in the parameter $a$. In particular, $a$ consists of four scalars: the factor for brightness enhancement in the range $(0.6, 1.4)$, the factor for contrast adjustment in the range $(0.6, 1.4)$, the factor for gamma transform in the range $(0.5,2.0)$ and the kernel size of the Gaussian blur in the range $(0.05,2.0)$. Essentially, the context is obtained by augmenting the original image \emph{twice}, where the second augmentation is modeled by \methodname{}. Domain variation modeling is implemented along with local or global anatomical modeling. The parameter $a$ is concatenated with the mask token $m\in \mathbb R^d$ along the feature dimension, resulting in a vector of length $d+4$, which is then fed into the policy network $\pi$. The policy network $\pi$ is a three-layer MLP with an input dimension of $d+4$ and an output dimension of $d$.

\subsection{Analytical Experiments}
\textbf{Anatomical modeling visualization.}
We utilize the RCDM framework \cite{bordes2021high} to showcase the anatomical modeling capabilities of \methodname{}. Specifically, we train a diffusion model to predict target pixel values conditioned on the output representation $\hat{h}_y$ of the world model. This guiding representation is first projected to a 512-dimensional vector, which is then injected into the diffusion model via conditional batch normalization layers \cite{dumoulin2016learned} within each residual block. 
For local anatomical structure modeling, the diffusion model individually predicts four rectangular masked regions, guided by spatially pooled predictor outputs corresponding to each location.
For global anatomical layout modeling, the model predicts the entire target region using spatially pooled outputs from the predictor. 
Figure 5 is built upon local anatomical modeling, focusing on masked regions with visible artifacts.
The diffusion model is trained using the validation split of the NIH ChestXray-14 dataset, while the visualizations are generated from the test split. This separation ensures that there is no information leakage between the different stages of the experiment—\methodname{} pre-training, diffusion model training, and visualization.

\textbf{Anatomical Correspondence Visualization.} We input the entire radiograph into the \methodname{} encoder to obtain image patch embeddings. Then we calculate per-pixel feature embeddings using RoI pooling over a 2x2 window centered on the pixel location. To illustrate anatomical correspondence, we focus on four key anatomical landmarks: the aortic arch, right hilum, left ventricle, and clavicle. The final similarity map is computed by measuring the L2 distance between the landmark embeddings of the reference image and the pixel embeddings of the test image. For improved visualization, the similarity values are rescaled.

\textbf{Domain sensitivity test.} To evaluate how effectively \methodname{} handles domain variations, we construct a test dataset using different augmentation configurations applied to the same base image. Specifically, we sample $n=64$ augmentation parameters evenly from a predefined range and apply these augmentations to generate a candidate set of target images $\{y_i\}_{i=1}^n$. For each target $y_i$, we further apply a randomly sampled augmentation to obtain the corresponding context $x_i=\mathcal T_{a_i}(y_i)$, resulting in a set of context-target-latent triplets $\{(x_i, y_i, a_i)\}$. The model's task is to predict the target $y_i$ given the context $x_i$ and latent $a_i$. The prediction error is defined as:
\begin{equation}
    L(y;x,a) = \|g(f_\theta(x); a) - f'_{\theta'}(y)\|^2.
\end{equation}
Ideally, the prediction error $L(y_i,x_i,a_i)$ should be smaller than $L(y_j,x_i,a_i)$ for any $j\ne i$. For the $i$-th case, we rank the errors $\{L(y_j,x_i,a_i)\}_{j=1}^n$ across the candidate set and compute the top-k recall rate of the true target $y_i$. This procedure is repeated across 50 different images, and the final result is the averaged outcome over these trials.

\subsection{Fine-tuning}
For classification, we employ mean pooling over all the output tokens to obtain a global feature representation of the image. Subsequently, a task-specific linear head is attached to the model for fine-tuning. We utilize the AdamW optimizer with a default learning rate of $1\times10^{-4}$, with layer-wise decay set to $0.75$ and a drop path rate of $0.6$. For the CheXpert benchmark, we adopt a learning rate of $1\times 10^{-2}$ and a drop path of $0.1$. The data augmentation pipeline involves random resized cropping and color jittering.

For segmentation, we connect a U-Net decoder with the pre-trained backbone with a SimpleFPN \cite{vitdet} adapter. The U-Net decoder has four stages with number of channels $8$, $16$, $32$, and $64$. The initial learning rate is set to $2\times10^{-4}$ with a layer-wise decay rate of $0.8$ and a drop path rate of $0.1$. The data preprocessing pipeline for training involves random brightness contrast, shifting, and scaling.

Due to the varying sizes of the datasets, we employ different batch sizes and epochs across benchmarks. The input size of the image is set to $224\times224$ pixels. 10\% of the training data is used for validation. Each experiment is conducted five times.

\section{Numerical Results}
Figure 7 illustrates the fine-tuning performance of \methodname{} on the VinDr-CXR dataset using varying proportions of the training data, which highlights \methodname{}'s ability to enhance data efficiency. Here, we present the corresponding numerical results in Table \ref{tab:result_frac}.

\begin{table}[h]
\centering
\caption{Fine-tuning with 1\%, 10\%, and 100\% training data on VinDr-CXR.}
\label{tab:result_frac}
\vspace{-5pt}
    \resizebox{0.9\linewidth}{!}{
    \begin{tabular}{c|ccc}
    \toprule
        Method & $1\%$ & $10\%$ & $100\%$ \\
        \midrule
        LVM-Med & 76.41$\pm$3.79&85.85$\pm$0.59 & 88.22$\pm$0.44 \\
        Adam-v2 & 77.90$\pm$1.14 & 88.26$\pm$0.48  & 91.46$\pm$0.33 \\
        MAE &78.07$\pm$1.66 & 90.63$\pm$0.16& 92.76$\pm$0.18\\
        SimMIM & 83.85$\pm$1.62&92.15$\pm$0.31& 92.81$\pm$0.31\\
        \midrule
        \methodname{} & \textbf{90.53$\pm$1.01} &\textbf{94.71$\pm$0.10} & \textbf{95.24$\pm$0.12}\\
        \bottomrule
    \end{tabular}}
\end{table}



\end{document}